%% file: Main.tex
\documentclass[letterpaper,conference]{IEEEtran}

\input{Format}

\begin{document}

\input{Title}
\input{Abstract}
\input{Introduction}
\input{RelatedWork}
\input{Method}
\input{Results}
\input{Conclusions}
\bibliographystyle{IEEEtran}
\bibliography{ref}

\end{document}

%% file: Format.tex
\IEEEoverridecommandlockouts

\usepackage{amsmath,amssymb,amsfonts}
\usepackage{array}
\usepackage[caption=false,font=normalsize,labelfont=sf,textfont=sf]{subfig}
\usepackage{textcomp}
\usepackage{stfloats}
\usepackage{url}
\usepackage{verbatim}
\usepackage{graphicx}
\usepackage{cite}
\usepackage{xspace}
\newcommand \name{PHR-VLA\xspace}

\usepackage[hidelinks]{hyperref}
\hypersetup{
   linkcolor=blue,
   breaklinks=true,   % splits links across lines
   colorlinks=true,   % displays links as colored text instead of blocks
   citecolor=blue,
   urlcolor=blue
}
\usepackage[table]{xcolor}
\usepackage{siunitx,booktabs}
\usepackage{multirow}
\usepackage{caption}
\usepackage{enumitem}
\usepackage{hhline}
\usepackage{comment}

\newcommand{\xxnote}[3]{}
\renewcommand{\xxnote}[3]{\color{#2}{#1: #3}}

\usepackage{censor}
\def\censorcolor{gray!50} \let\svcensorrule\censorrule \renewcommand\censorrule[1]{ \textcolor{\censorcolor}{\svcensorrule{#1}}}

\usepackage[ruled,vlined,linesnumbered]{algorithm2e}

\SetCommentSty{mycommfont}
\SetKwComment{Comment}{// }{}
\let\oldnl\nl%
\newcommand{\nonl}{\renewcommand{\nl}{\let\nl\oldnl}}%

\usepackage[switch]{lineno}
\usepackage[most]{tcolorbox}

\definecolor{phrvla}{HTML}{D43F3A}
\definecolor{baselines}{HTML}{7D3C98}

%% file: Title.tex
\title{PHR-VLA: Planning Horizon Reasoning for Vision-Language-Action Models}

\author{Davood Soleymanzadeh$^{1}$, Kaidi Zhang$^{2}$, Zhiyuan Zhang$^{2}$, Bihao Zhang$^{1}$, Xiao Liang$^{3}$, Yu She$^{2}$, Minghui Zheng$^{1}$
        % <-this % stops a space
\thanks{$^{1}$Davood Soleymanzadeh, Bihao Zhang, and Minghui Zheng are with the J. Mike Walker '66 Department of Mechanical Engineering, Texas A\&M University, College Station, TX 77843, USA (\tt\footnotesize e-mail: davoodso@tamu.edu; bhzhang@tamu.edu; mhzheng@tamu.edu).}% <-this % stops a space
\thanks{$^{2}$Kaidi Zhang, Zhiyuan Zhang, and Yu She are with the Department of Industrial Engineering, Purdue University, West Lafayette, IN 47907, USA (\tt\footnotesize e-mail: zhan5896@purdue.edu; zhan5570@purdue.edu; shey@purdue.edu).}
\thanks{$^{3}$Xiao Liang is with the Zachry Department of Civil and Environmental
Engineering, Texas A\&M University, College Station, TX 77843 USA (\tt\footnotesize e-mail:
xliang@tamu.edu).}
\thanks{This work was partially supported by the USA National Science Foundation under Grant No. 2527316, No. 2422826 and No. 2423068. Portions of this research were conducted with the advanced computing resources provided by Texas A\&M High Performance Research Computing.}
}

\makeatletter
\let\@oldmaketitle\@maketitle%
\renewcommand{\@maketitle}{\@oldmaketitle%
\setcounter{figure}{0}
\vspace{10pt}
\centering
\begin{center}
\includegraphics[width=0.9\linewidth]{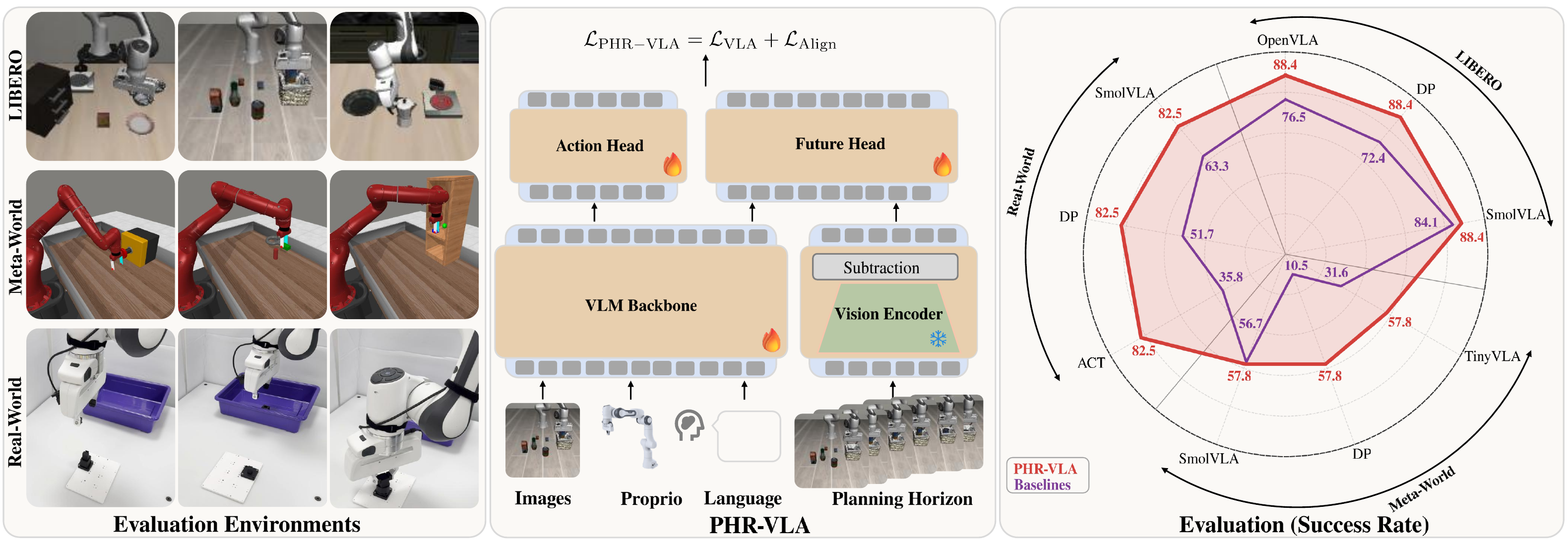}
  % \vspace{15pt}
  \captionof{figure}{\small \textbf{Method Overview.} \textit{Left:} The three evaluation domains: LIBERO~\cite{liu2023libero}, Meta-World~\cite{yu2020meta}, and real-world disassembly. \textit{Center:} \name fine-tunes a base VLA policy with an auxiliary future head that predicts planning-horizon latent dynamics from a frozen vision encoder, jointly optimizing $\mathcal{L}_{\mathrm{\name}} = \mathcal{L}_{\mathrm{VLA}} + \lambda \mathcal{L}_{\mathrm{Align}}$. The future head and vision encoder are used only during training. \textit{Right}: Across all three domains, \name (\textcolor{phrvla}{red}) consistently outperforms SmolVLA~\cite{shukor2025smolvla}, and other baseline policies (\textcolor{baselines}{purple}), including OpenVLA~\cite{kim2024openvla}, Diffusion Policy (DP)~\cite{chi2025diffusion}, TinyVLA~\cite{wen2025tinyvla}, and ACT~\cite{zhao2023learning}.}
  \label{fig:openning}
  \end{center}
  \vspace{-0.2in}
  }
\makeatother
\maketitle
\thispagestyle{empty}
\pagestyle{empty}

%% file: Abstract.tex
\begin{abstract}
Vision-language-action models (VLAs) have shown strong promise for general-purpose robotic manipulation by mapping language instructions and vision observations directly to actions. However, most VLAs primarily condition action prediction on current observations and lack an explicit mechanism for reasoning over future task dynamics, which is particularly important for fine-grained, contact-rich manipulation. We present \name, a framework that enables planning-horizon reasoning in VLAs through privileged latent representations of future dynamics. \name introduces a lightweight auxiliary future head that, during training, aligns the VLA's internal representations with latent dynamics extracted from future observations. Evaluation results demonstrate that local, contact-centric, patch-level latent dynamics supervision from the wrist camera improves success rate on LIBERO from \underline{84.1}\% to \underline{88.4}\% and on real-world disassembly tasks from \underline{63.3}\% to \underline{82.5}\%. Patch-level supervision from a third-person camera also improves performance on Meta-World from \underline{56.70}\% to \underline{57.8}\%. These results demonstrate that privileged latent dynamics alignment provides an effective training signal for improving anticipatory reasoning in VLA policies. Project website: \href{https://davoodsz.github.io/PHR-VLA.github.io/}{https://davoodsz.github.io/PHR-VLA.github.io/}
\end{abstract}


%% file: Introduction.tex
\section{Introduction} \label{sec: introduction}
Vision-language-action (VLA) models have demonstrated strong potential for general-purpose robotic manipulation~\cite{zitkovich2023rt, kim2024openvla, black2024pi_0, intelligence2025pi_}. By extending pretrained vision-language models (VLMs) to robot action prediction, VLAs ground language instructions and visual observations into executable control commands using large-scale robot demonstration datasets~\cite{o2024open}. Despite this progress, many VLAs remain largely reactive: they predict actions from current time-step observations without explicitly modeling how the task and scene may evolve over the planning horizon. This limitation can hinder temporally coordinated behavior in manipulation tasks that require anticipation of future dynamics~\cite{zhao2025cot}.

Recent studies have begun to address this limitation by incorporating memory and anticipatory reasoning into VLA policies. Memory-based approaches provide additional temporal context by conditioning the policy on a dense history of observations~\cite{torne2026mem, lee2024behavior, team2024octo}. However, processing dense, long-horizon observation histories can increase computational cost and inference latency, which is undesirable for real-time robotic control. To reduce this overhead, recent methods have explored compact memory representations, including proprioceptive states memory~\cite{zhang2025ta}, 2D point tracks~\cite{chen2025history}, selected keyframes~\cite{mark2026bpp}, and natural-language summaries~\cite{lin2025onetwovla}.

In parallel, forecasting-based approaches seek to improve embodied reasoning by predicting future-oriented representations before or during action generation. These representations include language descriptions, latent features, subgoal images, or visual foresight signals~\cite{bu2025agibot, intelligence2025pi_, zhao2025cot, zhang2026foreact}. Large video prediction and world-model-based methods further suggest that future prediction objectives can improve policy generalization to novel scenes and objects~\cite{li2026causal, ye2026world}. However, these approaches often require explicit future prediction, autoregressive reasoning, or additional inference modules. Such components can increase computational cost and complicate integration with pretrained VLAs for real-time control~\cite{zhang2026foreact}.

These observations motivate the following question: can a VLA learn future-aware planning representations without explicitly generating future observations or rollouts at inference time~\cite{yuan2026fast}? To address this question, we introduce \textbf{\name}, a framework for planning-horizon reasoning in VLA models through privileged latent dynamics representations. Rather than adding an explicit world model at deployment, \name uses latent representations of future observations as privileged supervision during training. Specifically, a lightweight auxiliary future head aligns current internal embeddings of the VLA with latent representations of future task dynamics over the planning horizon. This objective encourages the policy to encode information relevant to future task evolution while preserving the original action-generation pipeline at inference time.

Our main contributions are:
\begin{itemize}
    \item We introduce \name, a framework that leverages privileged latent dynamics to enable planning-horizon reasoning in VLAs. \name encourages the policy to encode future task dynamics without requiring explicit future rollouts or world-model inference during deployment.
    
    \item We systematically investigate different camera viewpoints, representation granularities, latent target formulations, and encoder choice for privileged future supervision. Our results show that short-horizon, local, and contact-centric signals are particularly effective. Patch-level wrist-camera supervision improves the success rate on LIBERO~\cite{liu2023libero} from \underline{\textbf{84.1}}\% to \underline{\textbf{88.4}}\% and on real-world disassembly tasks from \underline{\textbf{63.3}}\% to \underline{\textbf{82.5}}\%. Patch-level third-person-camera supervision also improves performance on Meta-World~\cite{yu2020meta} from \underline{\textbf{56.7}}\% to \underline{\textbf{57.8}}\%.
\end{itemize}

The remainder of this paper is organized as follows. Section~\ref{sec:relatedwork} reviews related work on VLAs, reasoning in robotic foundation models, and world models. Section~\ref{sec:phr-vla} presents the proposed \name framework. Section~\ref{sec:results} describes the experimental evaluation. Finally, Section~\ref{sec:conclusion} concludes the paper.

%% file: RelatedWork.tex
\section{Related Work} \label{sec:relatedwork}
We review prior work on vision-language-action models, temporal reasoning through memory and forecasting, and world models for robotic planning.

\vspace{0.2cm}
\noindent
\textbf{Vision-Language-Action Models (VLAs).} Vision-language-action models (VLAs) have emerged as a promising approach to general-purpose robotic manipulation by fine-tuning pretrained vision-language models (VLMs) on robot demonstration data~\cite{zitkovich2023rt, kim2024openvla, black2024pi_0, intelligence2025pi_, o2024open, bjorck2025gr00t}. These models use either discrete action decoders~\cite{kim2024openvla} or continuous policy heads~\cite{black2024pi_0} to generate robot actions. By leveraging the semantic knowledge acquired during large-scale VLM pretraining, VLAs can generalize across tasks, objects, and environments. However, most VLAs primarily generate actions from the current time-step visual language context, with limited explicit representation of how the task may evolve over the planning horizon~\cite{zheng2025flare, fang2026molmoact2}. In contrast, \name improves the internal planning horizon representations of a VLA while preserving its original action-generation interface.

\vspace{0.2cm}
\noindent
\textbf{Improving VLAs with Memory and Forecasting.} A growing body of work has investigated the use of temporal context to improve VLA policies~\cite{torne2026mem, zheng2025tracevla}. One line of research conditions the policy on a dense history of observations~\cite{lee2024behavior, team2024octo}. Although dense observation histories provide useful temporal information, they also increase input sequence length, computational cost, and inference latency~\cite{torne2026mem}. Recent methods have therefore explored more compact memory representations, including proprioceptive state memory~\cite{zhang2025ta}, 2D point tracks~\cite{chen2025history}, selected keyframes~\cite{mark2026bpp}, and natural-language summaries~\cite{lin2025onetwovla}. 

Another line of research introduces explicit reasoning or forecasting stages before action generation. These methods generate intermediate representations such as language reasoning traces, feature embeddings, subgoal images, visual foresight signals, or future action representations~\cite{bu2025agibot, intelligence2025pi_, zhao2025cot, zhang2026foreact, intelligence2025pi1, intelligence2026pi2}. Such representations can improve task-level planning and temporal action consistency by encouraging the policy to reason beyond the current observation. However, many of these methods require additional inference modules, autoregressive reasoning steps, or explicit prediction stages, which can increase latency and complicate their integration with pretrained VLA architectures. \name leverages privileged future dynamics reasoning only during training and does not require explicit future rollouts or additional reasoning stages during deployment.

\vspace{0.2cm}
\noindent
\textbf{World Models in Robotics.} World models and video prediction models support future-aware decision-making by explicitly modeling how the environment evolves over time~\cite{li2026causal, ye2026world, guo2026vlaw}. By learning predictive latent representations or generating future visual rollouts, these methods can support planning, policy learning, and generalization to novel scenes and object configurations. Recent studies have shown that coupling future prediction objectives with action prediction can improve the generalization of robotic policies~\cite{li2026causal, ye2026world}.

However, world-model- and video-prediction-based policies often rely on future rollout generation, autoregressive latent prediction, or large generative architectures at inference time~\cite{zhang2026foreact}. These components can introduce substantial computational cost and latency, particularly in real-time manipulation settings. \name draws on the representational benefits of future dynamics modeling while avoiding explicit world-model inference during deployment~\cite{yuan2026fast}. Rather than generating future observations or rollouts online, \name uses a lightweight alignment objective to transfer privileged future dynamics information into the internal representations of the VLA during training. Therefore, \name complements world-model-based approaches by improving future-aware action generation while maintaining the inference pipeline and computational efficiency of the underlying VLA.

%% file: Method.tex
\section{\name} \label{sec:phr-vla}
We introduce \name, a training-time auxiliary supervision framework that uses privileged future latent dynamics to improve planning horizon reasoning in VLAs.

\begin{figure*}[t]
    \centering
    \includegraphics[width=\textwidth]{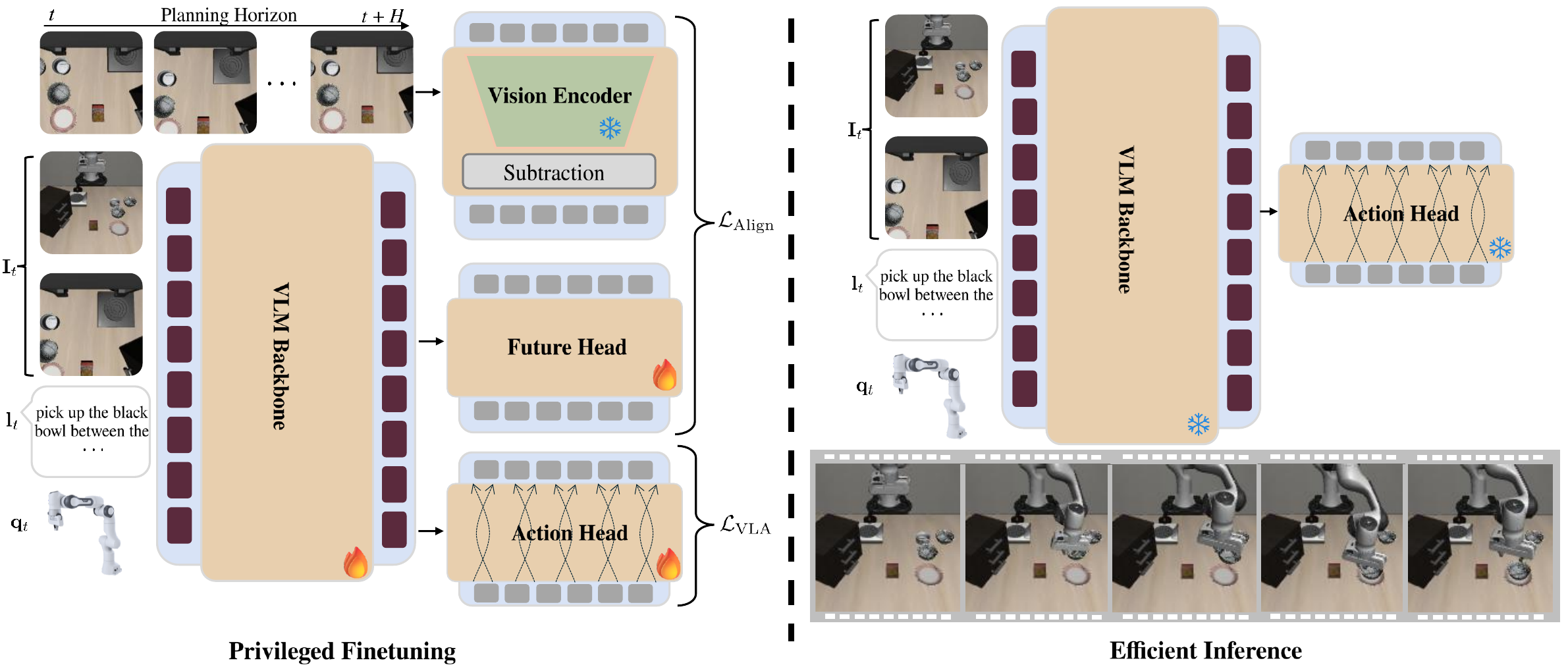}
    \captionof{figure}{\small \textbf{\name Framework.} \textbf{Privileged Finetuning:} In addition to the current observation $\mathbf{I}_t$, language instruction $\mathbf{l}_t$, and proprioceptive state $\mathbf{q}_t$ used by the standard VLA backbone, privileged planning horizon frames $\mathbf{I}_{t+1:t+H}$ are passed through a frozen vision encoder to compute latent-dynamics targets. The action-token latents produced by VLM backbone feed two heads: the standard flow matching action head, supervised by $\mathcal{L}_{\mathrm{VLA}}$, and the auxiliary future head which is supervised by $\mathcal{L}_{\mathrm{Align}}$ against the latent dynamics targets. \textbf{Efficient Inference:} The action head runs exactly as in standard VLA backbone, so \name adds zero latency or memory cost at deployment.}
    \label{fig:framework}
\end{figure*}

\subsection{VLAs}
A standard VLA framework receives an observation $\mathbf{o}_t \triangleq (\mathbf{I}_t, \mathbf{l}_t, \mathbf{q}_t)$ at each timestep $t$ and predicts an action chunk $\mathbf{a}_{t:t+H}$ over a planning horizon $H$. Here, $\mathbf{I}_t = \{I_t^i\}_{i=1}^N$ denotes a set of $N$ camera images, $\mathbf{l}_t$ is the language instruction, and $\mathbf{q}_t$ is the robot proprioceptive state. The visual observations and proprioceptive state are encoded using modality-specific encoders and projected into the token embedding space of the VLM backbone, together with the tokenized language instruction. The backbone produces multimodal representations, which are processed by an action head to parameterize an action distribution~\cite{xu2026rl}. The model is trained on a demonstration dataset $\mathcal{D}$ with the standard log-likelihood objective:
\begin{equation}
 \max_{\theta} \ \mathbb{E}_{(\mathbf{a}_{t:t+H}, \mathbf{o}_t)\sim \mathcal{D}} \left[\log p_{\theta}(\mathbf{a}_{t:t+H} \mid \mathbf{o}_t)\right].
\end{equation}

\subsection{VLA Baseline}
We adopt SmolVLA (0.45B)~\cite{shukor2025smolvla} as our VLA baseline. SmolVLA couples a compact SmolVLM-2 backbone with a flow-matching action expert that predicts continuous action chunks given visual, language, and proprioceptive inputs. We choose SmolVLA because it provides strong VLA representations while remaining computationally efficient. We can fine-tune it on a single A100 GPU, which enables controlled ablations of our planning horizon dynamics-aware alignment.

\subsection{Planning Horizon Latent Dynamics}
\name uses future planning horizon observations available in the demonstration dataset as privileged training information. We encode the observations over the planning horizon $\mathbf{o}_{t:t+H}$ using a frozen visual encoder $\mathcal{E}_{\mathrm{encoder}}$~\cite{zhai2023sigmoid, assran2025v}. Let
\begin{equation}
\mathbf{s}_{t:t+H}^c = \mathcal{E}_{\mathrm{encoder}}(\mathbf{I}_{t:t+H}^c)
\end{equation}
denote the visual latents of the planning horizon observation, where $c$ denotes the camera used for the auxiliary target. Rather than supervising the policy to match absolute planning horizon visual latents, the main formulation supervises planning horizon latent dynamics:
\begin{equation}
\mathbf{y}_{t:t+H}^c = \mathbf{s}_{t+1:t+H}^c - \mathbf{s}_{t:t+H-1}^c.
\end{equation}

This target emphasizes how the scene representation changes over the planning horizon, while reducing the need to predict static visual content already present in the current timestep observation.

\begin{table}[htbp]
    \centering
    \captionof{table}{\small \textbf{Results on LIBERO.} Success rates are reported across four task suites (300 trials per suite). Entries marked with \textbf{$^*$} are taken from~\cite{kim2025fine} and may use different model sclaes, training setups, and computational budgets.  The rest are trained under the same a single A100 GPU setting. Best results are \textbf{bold}, and second-best results are \underline{underlined}.}
    \label{tab:libero-main}
    \resizebox{0.49\textwidth}{!}{%
    \begin{tabular}{lccccc}
        \toprule
         \rowcolor{gray!20} \multicolumn{1}{c|}{\textbf{Method}}&\multicolumn{1}{c}{\textbf{Spatial} $\uparrow$}&\multicolumn{1}{c}{\textbf{Object} $\uparrow$}&\multicolumn{1}{c}{\textbf{Goal} $\uparrow$}&\multicolumn{1}{c}{\textbf{Long} $\uparrow$}&\multicolumn{1}{|c}{\textbf{Average} $\uparrow$}  \\
         \midrule
         \multicolumn{1}{c|}{\textbf{ACT~\cite{zhao2023learning}}}&\multicolumn{1}{c}{30.7\%}&\multicolumn{1}{c}{48.4\%}&\multicolumn{1}{c}{02.0\%}&\multicolumn{1}{c}{15.7\%}&\multicolumn{1}{|c}{24.2\%} \\
         \midrule
         \multicolumn{1}{c|}{\textbf{Diffusion Policy~\cite{chi2025diffusion}$^*$}}&\multicolumn{1}{c}{78.3\%}&\multicolumn{1}{c}{92.5\%}&\multicolumn{1}{c}{68.3\%}&\multicolumn{1}{c}{50.5\%}&\multicolumn{1}{|c}{72.4\%} \\
         \multicolumn{1}{c|}{\textbf{Octo~\cite{team2024octo}$^*$}}&\multicolumn{1}{c}{78.9\%}&\multicolumn{1}{c}{85.7\%}&\multicolumn{1}{c}{84.6\%}&\multicolumn{1}{c}{51.1\%}&\multicolumn{1}{|c}{75.1\%} \\
         \multicolumn{1}{c|}{\textbf{DiT Policy~\cite{hou2024diffusion}$^*$}}&\multicolumn{1}{c}{84.2\%}&\multicolumn{1}{c}{96.3\%}&\multicolumn{1}{c}{85.4\%}&\multicolumn{1}{c}{63.8\%}&\multicolumn{1}{|c}{82.4\%} \\
        \multicolumn{1}{c|}{\textbf{OpenVLA~\cite{kim2024openvla}$^*$}}&\multicolumn{1}{c}{84.7\%}&\multicolumn{1}{c}{88.4\%}&\multicolumn{1}{c}{79.2\%}&\multicolumn{1}{c}{53.7\%}&\multicolumn{1}{|c}{76.5\%} \\
        \midrule
        \multicolumn{1}{c|}{\textbf{SmolVLA~\cite{shukor2025smolvla}}}&\multicolumn{1}{c}{\underline{86.0}\%}&\multicolumn{1}{c}{\underline{97.7}\%}&\multicolumn{1}{c}{\underline{86.7}\%}&\multicolumn{1}{c}{\underline{66.0}\%}&\multicolumn{1}{|c}{\underline{84.1}\%} \\
        \midrule
        \multicolumn{1}{c|}{\textbf{\name (ours)}}&\multicolumn{1}{c}{\textbf{88.7}\%}&\multicolumn{1}{c}{\textbf{98.0}\%}&\multicolumn{1}{c}{\textbf{92.4}\%}&\multicolumn{1}{c}{\textbf{74.4}\%}&\multicolumn{1}{|c}{\textbf{88.4}\%} \\
        \bottomrule
    \end{tabular}}
\end{table}

\subsection{Planning Horizon Reasoning}
Given the current observation and language instruction, SmolVLA~\cite{shukor2025smolvla} produces an action chunk together with an internal action-token representation $\mathbf{z}_t$. \name attaches a lightweight future head $g_\phi$ that predicts planning horizon latent dynamics from these action-token representations:
\begin{equation}
\hat{\mathbf{y}}_{t,t+H}^c = g_{\phi}(\mathbf{z}_t).
\end{equation}

The head encourages the planned-action representation to encode how the visual scene is expected to evolve across the action horizon. The predicted and privileged latent dynamics are aligned alongside the standard VLA training objective as follows:
\begin{equation}
\mathcal{L}_{\mathrm{\name}} = \mathcal{L}_{\mathrm{VLA}} + \lambda \mathcal{L}_{\mathrm{Align}}, 
\end{equation}
where the alignment loss $\mathcal{L}_{\mathrm{Align}}$ is computed as the mean-squared error between predicted and privileged planning-horizon latent dynamics and $\lambda$ controls the strength of planning horizon dynamics supervision.

\begin{table}[htbp]
    \centering
    \captionof{table}{\small \textbf{Results on Meta-World.} Success rates are reported across all difficulty levels. Entries marked with \textbf{$^*$} are taken from~\cite{shukor2025smolvla}and may use different model sclaes, training setups, and computational budgets.  The rest are trained under the same a single A100 GPU setting. Best results are \textbf{bold}, and second-best results are \underline{underlined}.}
    \label{tab:metaworld-main}
    \resizebox{0.49\textwidth}{!}{%
    \begin{tabular}{lccccc}
        \toprule
         \rowcolor{gray!20} \multicolumn{1}{c|}{\textbf{Method}}&\multicolumn{1}{c}{\textbf{Easy} $\uparrow$}&\multicolumn{1}{c}{\textbf{Medium} $\uparrow$}&\multicolumn{1}{c}{\textbf{Hard} $\uparrow$}&\multicolumn{1}{c}{\textbf{VeryHard} $\uparrow$}&\multicolumn{1}{|c}{\textbf{Average} $\uparrow$}  \\
         \midrule
         \multicolumn{1}{c|}{\textbf{ACT~\cite{zhao2023learning}}}&\multicolumn{1}{c}{68.2\%}&\multicolumn{1}{c}{49.7\%}&\multicolumn{1}{c}{28.9\%}&\multicolumn{1}{c}{49.4\%}&\multicolumn{1}{|c}{49.0\%} \\
         \midrule
         \multicolumn{1}{c|}{\textbf{Diffusion Policy~\cite{chi2025diffusion}$^*$}}&\multicolumn{1}{c}{23.1\%}&\multicolumn{1}{c}{10.7\%}&\multicolumn{1}{c}{01.9\%}&\multicolumn{1}{c}{06.1\%}&\multicolumn{1}{|c}{10.5\%} \\
         \multicolumn{1}{c|}{\textbf{TinyVLA~\cite{wen2025tinyvla}$^*$}}&\multicolumn{1}{c}{77.6\%}&\multicolumn{1}{c}{21.5\%}&\multicolumn{1}{c}{11.4\%}&\multicolumn{1}{c}{15.8\%}&\multicolumn{1}{|c}{31.6\%} \\
        \midrule
        \multicolumn{1}{c|}{\textbf{SmolVLA~\cite{shukor2025smolvla}}}&\multicolumn{1}{c}{\underline{82.0}\%}&\multicolumn{1}{c}{\underline{54.2}\%}&\multicolumn{1}{c}{\underline{43.9}\%}&\multicolumn{1}{c}{\textbf{46.7}\%}&\multicolumn{1}{|c}{\underline{56.7}\%} \\
        \midrule
        \multicolumn{1}{c|}{\textbf{\name (ours)}}&\multicolumn{1}{c}{\textbf{85.2}\%}&\multicolumn{1}{c}{\textbf{55.1}\%}&\multicolumn{1}{c}{\textbf{45.5}\%}&\multicolumn{1}{c}{\underline{45.3}\%}&\multicolumn{1}{|c}{\textbf{57.7}\%} \\
        \bottomrule
    \end{tabular}}
\end{table}

\subsection{Inference}
\name uses future observations over the planning horizon and the auxiliary future head during training. At inference time, the future head is discarded, and the policy executes the original SmolVLA~\cite{shukor2025smolvla} action given current timestep observations.

%% file: Results.tex
\section{Results and Discussion} \label{sec:results}
We evaluate whether \name improves VLA fine-tuning by supervising the policy with planning-horizon latent scene dynamics. Specifically, we address the following questions: (1) Does privileged future-latent dynamics supervision improve standard VLA fine-tuning? (2) Is supervising changes in the visual latent space more effective than supervising absolute future visual embeddings? (3) Which camera view is most effective for planning-horizon reasoning?  

\begin{figure}[htbp]
    \includegraphics[width=0.9\linewidth]{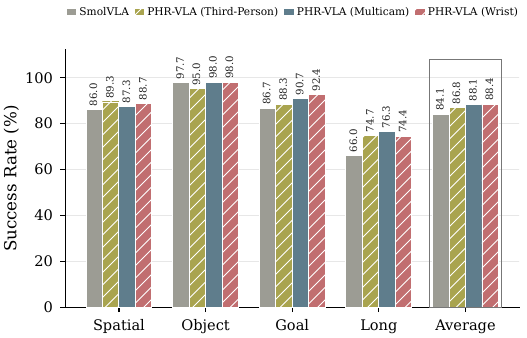}
    \centering
    \captionof{figure}{\small \textbf{Ablation: Camera View.} Success rate of SmolVLA and \name with wrist-mounted, fixed third-person, and multi-camera auxiliary supervision.}
    \label{fig:ablation_cameraview}
\end{figure}

\subsection{Baselines \& Evaluation Protocol}
\noindent
\textbf{Baselines.} We compare \name against recent VLA and imitation-learning baselines: ACT~\cite{zhao2023learning}, Diffusion Policy~\cite{chi2025diffusion}, Octo~\cite{team2024octo}, DiT Policy~\cite{hou2024diffusion}, OpenVLA~\cite{kim2024openvla}, TinyVLA~\cite{wen2025tinyvla}, and SmolVLA~\cite{shukor2025smolvla}.

\begin{figure}[htbp]
    \includegraphics[width=0.9\linewidth]{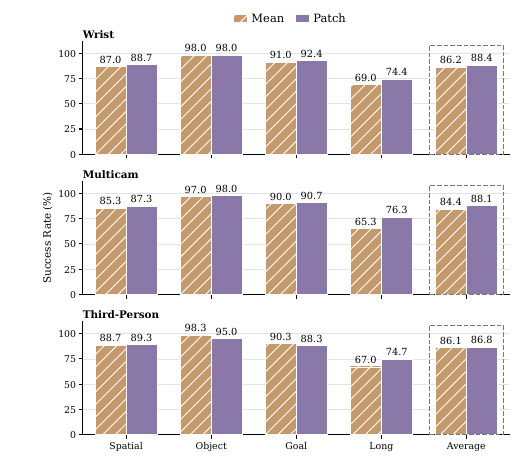}
    \centering
    \captionof{figure}{\small \textbf{Ablation: Target Granularity.} Success rate of \name with various target supervision granularity across wrist-mounted, fixed-third person, and multi-camera.}
    \label{fig:ablation_targetgarnularity}
\end{figure}

\vspace{0.2cm}
\noindent
\textbf{Evaluation Protocol.} We evaluate \name on two multi-task simulation benchmarks: LIBERO~\cite{liu2023libero}, and Meta-World~\cite{yu2020meta}. LIBERO is a language-conditioned tabletop manipulation benchmark that evaluates multi-task generalization across task families with varying spatial layouts, target goals, and object configurations. Meta-World is a simulated robotic manipulation benchmark that evaluates generalization across diverse manipulation tasks with varying object configurations and goal conditions.

\begin{figure}[htbp]
    \includegraphics[width=0.9\linewidth]{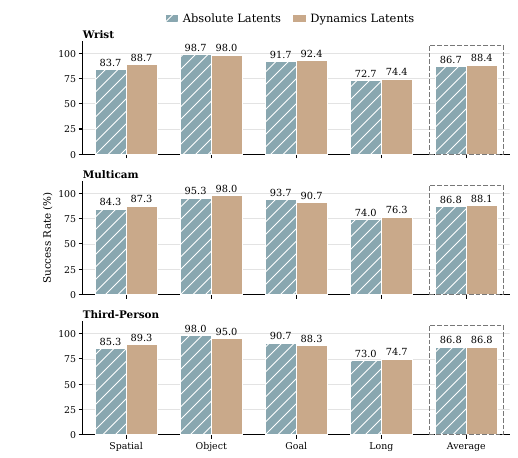}
    \centering
    \captionof{figure}{\small \textbf{Ablation: Absolute Latent vs. Dynamics Latent Supervision.} Success rate of \name with planning-horizon absolute latents and dynamics latent supervision during training.}
    \label{fig:ablation_absolutevslatent}
\end{figure}

\begin{figure}[htbp]
    \includegraphics[width=0.9\linewidth]{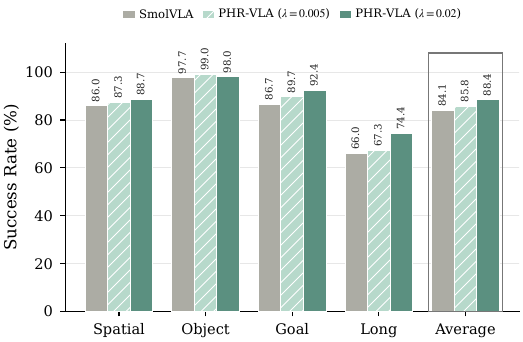}
    \centering
    \captionof{figure}{\small \textbf{Ablation: $\lambda$.} Success rate of SmolVLA and \name with various auxiliary loss weights ($\lambda$).}
    \label{fig:ablation_lambda}
\end{figure}

\subsection{Results} \label{subsec:mainresults}
\noindent
\textbf{LIBERO.} As shown in Table~\ref{tab:libero-main}, under the controlled single-A100 fine-tuning setting, \name consistently improves the SmolVLA baseline across all four LIBERO suites.. More specifically, \name improves standard SmolVLA fine-tuning when the auxiliary objective supervises latent scene dynamics over the planning horizon. \name increases the average LIBERO success rate from \underline{84.1}\% to \underline{88.4}\%. The gains are most pronounced on LIBERO-Long, where \name improves performance by +8 points. This suggests that future-latent fine-grained dynamics supervision through wrist camera is especially useful in settings where the policy's success depends on long-horizon reasoning and manipulation.

\begin{table}[t]
\centering
\captionof{table}{\small \textbf{Ablation: Encoder Choice, SigLIP vs. JEPA on LIBERO.} Success rates are reported across four task suites (300 trials per task) for SigLIP and JEPA encoders auxiliary supervision. ``$\mathbf{y}$'' denotes planning-horizon latent dynamics supervision, and ``$\mathbf{s}$'' denotes planning-horizon absolute latent supervision. Best results are \textbf{bold}, and second-best results are \underline{underlined}.}
\label{tab:ablation_siglip_vs_jepa_libero}
\resizebox{0.49\textwidth}{!}{%
\begin{tabular}{llll|cccc|c}
\toprule
\rowcolor{gray!20} \textbf{Method} & \textbf{View} & \textbf{Target} & \textbf{Type} & \textbf{Spatial} $\uparrow$ & \textbf{Object} $\uparrow$ & \textbf{Goal} $\uparrow$ & \textbf{Long} $\uparrow$ & \textbf{Average} $\uparrow$ \\
\midrule
\textbf{smolVLA~\cite{shukor2025smolvla}} & -- & -- & -- & 86.0\% & 97.7\% & 86.7\% & 66.0\% & 84.1\% \\
\midrule
\multirow{12}{*}{\shortstack{\textbf{\name} \\ \textbf{(SigLIP)}}}
 & \textbf{Wrist} & \textbf{Patch} & $\mathbf{y}$        & 88.7\% & 98.0\% & 92.4\% & 74.4\% & \textbf{88.4}\% \\
 & \textbf{Wrist} & \textbf{Patch} & $\mathbf{s}$        & 83.7\% & \textbf{98.7}\% & 91.7\% & 72.7\% & 86.7\% \\
 & \textbf{Wrist} & \textbf{Mean}  & $\mathbf{y}$        & 87.0\% & 98.0\% & 91.0\% & 69.0\% & 86.2\% \\
 & \textbf{Wrist} & \textbf{Mean}  & $\mathbf{s}$        & 84.0\% & 98.0\% & 89.0\% & 63.3\% & 83.6\% \\
 & \textbf{Multicam} & \textbf{Patch} & $\mathbf{y}$     & 87.3\% & 98.0\% & 90.7\% & \textbf{76.3}\% & \underline{88.1}\% \\
 & \textbf{Multicam} & \textbf{Patch} & $\mathbf{s}$     & 84.3\% & 95.3\% & \textbf{93.7}\% & 74.0\% & 86.8\% \\
 & \textbf{Multicam} & \textbf{Mean}  & $\mathbf{y}$     & 85.3\% & 97.0\% & 90.0\% & 65.3\% & 84.4\% \\
 & \textbf{Multicam} & \textbf{Mean}  & $\mathbf{s}$     & 87.0\% & 97.0\% & 91.0\% & 65.7\% & 85.2\% \\
 & \textbf{Third-Person} & \textbf{Patch} & $\mathbf{y}$ & 89.3\% & 95.0\% & 88.3\% & \underline{74.7}\% & 86.8\% \\
 & \textbf{Third-Person} & \textbf{Patch} & $\mathbf{s}$ & 85.3\% & 98.0\% & 90.7\% & 73.0\% & 86.8\% \\
 & \textbf{Third-Person} & \textbf{Mean}  & $\mathbf{y}$ & 88.7\% & \underline{98.3}\% & 90.3\% & 67.0\% & 86.1\% \\
 & \textbf{Third-Person} & \textbf{Mean}  & $\mathbf{s}$ & 84.0\% & 96.3\% & 90.0\% & 71.7\% & 85.5\% \\
\midrule
\multirow{12}{*}{\shortstack{\textbf{\name} \\ \textbf{(JEPA)}}}
 & \textbf{Wrist} & \textbf{Patch} & $\mathbf{y}$        & 87.3\% & 95.7\% & 91.3\% & 70.7\% & 86.2\% \\
 & \textbf{Wrist} & \textbf{Patch} & $\mathbf{s}$       & 83.7\% & 97.3\% & 88.0\% & 69.0\% & 84.5\% \\
 & \textbf{Wrist} & \textbf{Mean}  & $\mathbf{y}$        & 82.0\% & 95.3\% & 89.0\% & 64.7\% & 82.8\% \\
 & \textbf{Wrist} & \textbf{Mean}  & $\mathbf{s}$        & 88.7\% & 94.7\% & 88.7\% & 66.0\% & 84.5\% \\
 & \textbf{Multicam} & \textbf{Patch} & $\mathbf{y}$     & 87.3\% & 95.7\% & 90.0\% & 70.7\% & 85.9\% \\
 & \textbf{Multicam} & \textbf{Patch} & $\mathbf{s}$     & 89.0\% & 95.0\% & 85.3\% & 67.7\% & 84.2\% \\
 & \textbf{Multicam} & \textbf{Mean}  & $\mathbf{y}$     & 89.3\% & 95.7\% & 92.0\% & 65.3\% & 85.6\% \\
 & \textbf{Multicam} & \textbf{Mean}  & $\mathbf{s}$     & \textbf{91.7}\% & 95.3\% & 92.0\% & 63.0\% & 85.5\% \\
 & \textbf{Third-Person} & \textbf{Patch} & $\mathbf{y}$ & 85.0\% & 95.7\% & 87.7\% & 64.3\% & 83.2\% \\
 & \textbf{Third-Person} & \textbf{Patch} & $\mathbf{s}$ & 87.7\% & 95.0\% & 87.7\% & 63.0\% & 83.3\% \\
 & \textbf{Third-Person} & \textbf{Mean}  & $\mathbf{y}$ & 89.3\% & 95.3\% & 90.7\% & 62.3\% & 84.4\% \\
 & \textbf{Third-Person} & \textbf{Mean}  & $\mathbf{s}$ & \underline{90.3}\% & 95.3\% & \underline{92.7}\% & 63.7\% & 85.5\% \\
\bottomrule
\end{tabular}}
\end{table}

\begin{figure}[htbp]
    \includegraphics[width=0.9\linewidth]{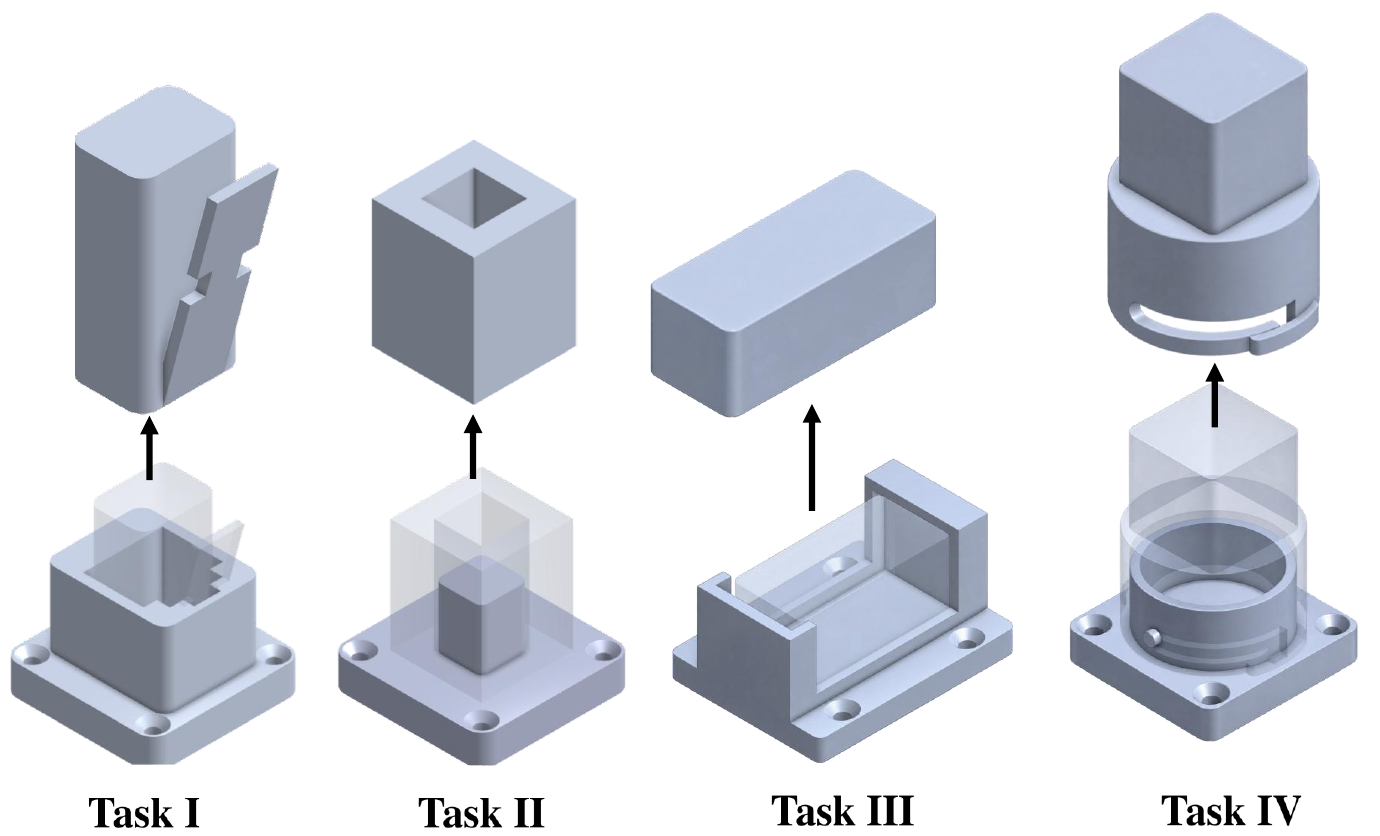}
    \centering
    \captionof{figure}{\small \textbf{Disassembly Tasks.} Contact-rich disassembly tasks for real-world evaluation of \name and baseline policies.}
    \label{fig:disassembly_tasks}
\end{figure}

\begin{table}[t]
\centering
\captionof{table}{\small \textbf{Ablation: Encoder Choice, SigLIP vs. JEPA on Meta-World.} Success rates are reported across all difficulty levels for SigLIP and JEPA encoders auxiliary supervision. ``$\mathbf{y}$'' denotes planning-horizon latent dynamics supervision, and ``$\mathbf{s}$'' denotes planning-horizon absolute latent supervision. Best results are \textbf{bold}, and second-best results are \underline{underlined}.}
\label{tab:ablation_siglip_vs_jepa_metaworld}
\resizebox{0.49\textwidth}{!}{%
\begin{tabular}{lll|cccc|c}
\toprule
\rowcolor{gray!20} \textbf{Method} & \textbf{Target} & \textbf{Type} & \textbf{Easy} $\uparrow$& \textbf{Medium} $\uparrow$& \textbf{Hard} $\uparrow$& \textbf{Very Hard} $\uparrow$& \textbf{Average} $\uparrow$ \\
\midrule
\textbf{smolVLA~\cite{shukor2025smolvla}} & -- & -- & 82.0\% & 54.2\% & 43.9\% & 46.7\% & 56.7\% \\
\midrule
\multirow{4}{*}{\shortstack{\textbf{\name} \\ \textbf{(SigLIP)}}}
 & \textbf{Patch} & $\mathbf{y}$    & 85.2\% & 55.1\% & 45.5\% & 45.3\% & 57.8\% \\
 & \textbf{Patch} & $\mathbf{s}$    & 84.2\% & \textbf{57.6}\% & 38.3\% & 47.3\% & 56.9\% \\
 & \textbf{Mean}  & $\mathbf{y}$    & 84.3\% & 55.8\% & 50.0\% & 44.7\% & 58.7\% \\
 & \textbf{Mean}  & $\mathbf{s}$    & 84.2\% & \textbf{57.6}\% & 38.3\% & 47.3\% & 56.9\% \\
\midrule
\multirow{4}{*}{\shortstack{\textbf{\name} \\ \textbf{(JEPA)}}}
 & \textbf{Patch} & $\mathbf{y}$    & 82.0\% & 51.5\% & 48.3\% & \underline{48.0}\% & 57.5\% \\
 & \textbf{Patch} & $\mathbf{s}$    & \underline{85.2}\% & \underline{56.1}\% & 43.3\% & \textbf{50.0}\% & 58.7\% \\
 & \textbf{Mean}  & $\mathbf{y}$    & \textbf{85.7}\% & 53.9\% & \underline{51.7}\% & \underline{48.0}\% & \textbf{59.8}\% \\
 & \textbf{Mean}  & $\mathbf{s}$    & 85.1\% & 53.0\% & \textbf{53.3}\% & 46.7\% & \underline{59.5}\% \\
\bottomrule
\end{tabular}}
\end{table}

\vspace{0.2cm}
\noindent
\textbf{Meta-World.} As shown in Table~\ref{tab:metaworld-main}, \name improves the overall average Meta-World success rate compared to the considered baseline. More specifically, \name improves standard SmolVLA fine-tuning when the auxiliary objective supervises latent scene dynamics over the planning horizon. However, given that Meta-World simulation environment only has third-view camera, the planning-horizon reasoning through this camera cannot provide fine-grained supervision during training.

\setlength{\tabcolsep}{3pt}
\begin{table*}[t]
\begin{center}
\captionof{table}{\small \textbf{Real-world Deployment.} Success rate of \name and baseline methods on real-world disassembly tasks. Best results are \textbf{bold}, and second-best results are \underline{underlined}.}
\label{tab:realworlddeployment}
\resizebox{0.9\textwidth}{!}{%
\begin{tabular}
{>{\raggedright\arraybackslash}p{0.07\textwidth}>{\raggedright\arraybackslash}p{0.33\textwidth}>{\raggedright\arraybackslash}p{0.07\textwidth}>{\raggedright\arraybackslash}p{0.14\textwidth}>{\raggedright\arraybackslash}p{0.1\textwidth}>{\raggedright\arraybackslash}p{0.08\textwidth}>{\raggedright\arraybackslash}p{0.13\textwidth}}
\toprule
\rowcolor{gray!20} \scriptsize{\textbf{Task}}&\scriptsize{\textbf{Prompt}}&\multicolumn{1}{c}{\scriptsize{\textbf{ACT~\cite{zhao2023learning}}}}&\multicolumn{1}{c}{\scriptsize{\textbf{Diffusion Policy~\cite{chi2025diffusion}}}}&\multicolumn{1}{c}{\scriptsize{\textbf{SmolVLA~\cite{shukor2025smolvla}}}}&\multicolumn{1}{c}{\scriptsize{\textbf{\name}}}&\multicolumn{1}{c}{\scriptsize{\textbf{\name (JEPA)}}} \\
\midrule
\scriptsize{\textbf{Task I}}&Press the clip to release the object, remove it, then place it in the purple container.&\multicolumn{1}{c}{19/30}&\multicolumn{1}{c}{21/30}&\multicolumn{1}{c}{26/30}&\multicolumn{1}{c}{\underline{28/30}}&\multicolumn{1}{c}{\textbf{29/30}} \\
\midrule
\scriptsize{\textbf{Task II}}&Remove the object from the loose shaft, then place it in the purple container.&\multicolumn{1}{c}{14/30}&\multicolumn{1}{c}{19/30}&\multicolumn{1}{c}{20/30}&\multicolumn{1}{c}{\underline{26/30}}&\multicolumn{1}{c}{\textbf{29/30}} \\
\midrule
\scriptsize{\textbf{Task III}}&Slide the object inward, pull it out, then place it in the purple container.&\multicolumn{1}{c}{10/30}&\multicolumn{1}{c}{14/30}&\multicolumn{1}{c}{\underline{24/30}}&\multicolumn{1}{c}{22/30}&\multicolumn{1}{c}{\textbf{26/30}}\\
\midrule
\scriptsize{\textbf{Task IV}}&Twist the object 90 degrees and pull it out from the shaft, then place it in the purple container.&\multicolumn{1}{c}{00/30}&\multicolumn{1}{c}{08/30}&\multicolumn{1}{c}{06/30}&\multicolumn{1}{c}{\textbf{23/30}}&\multicolumn{1}{c}{\underline{09/30}} \\
\midrule
\scriptsize{\textbf{Total [\#] $\uparrow$}}&\multicolumn{1}{c}{--}&\multicolumn{1}{c}{43/120}&\multicolumn{1}{c}{62/120}&\multicolumn{1}{c}{76/120}&\multicolumn{1}{c}{\textbf{99/120}}&\multicolumn{1}{c}{\underline{93/120}} \\
\midrule
\scriptsize{\textbf{Total [\%] $\uparrow$}}&\multicolumn{1}{c}{--}&\multicolumn{1}{c}{35.8\%}&\multicolumn{1}{c}{51.7\%}&\multicolumn{1}{c}{63.3\%}&\multicolumn{1}{c}{\textbf{82.5\%}}&\multicolumn{1}{c}{\underline{77.5\%}} \\
\bottomrule
\end{tabular}}
\end{center}
\end{table*}

\begin{figure*}[htbp]
    \includegraphics[width=0.85\linewidth]{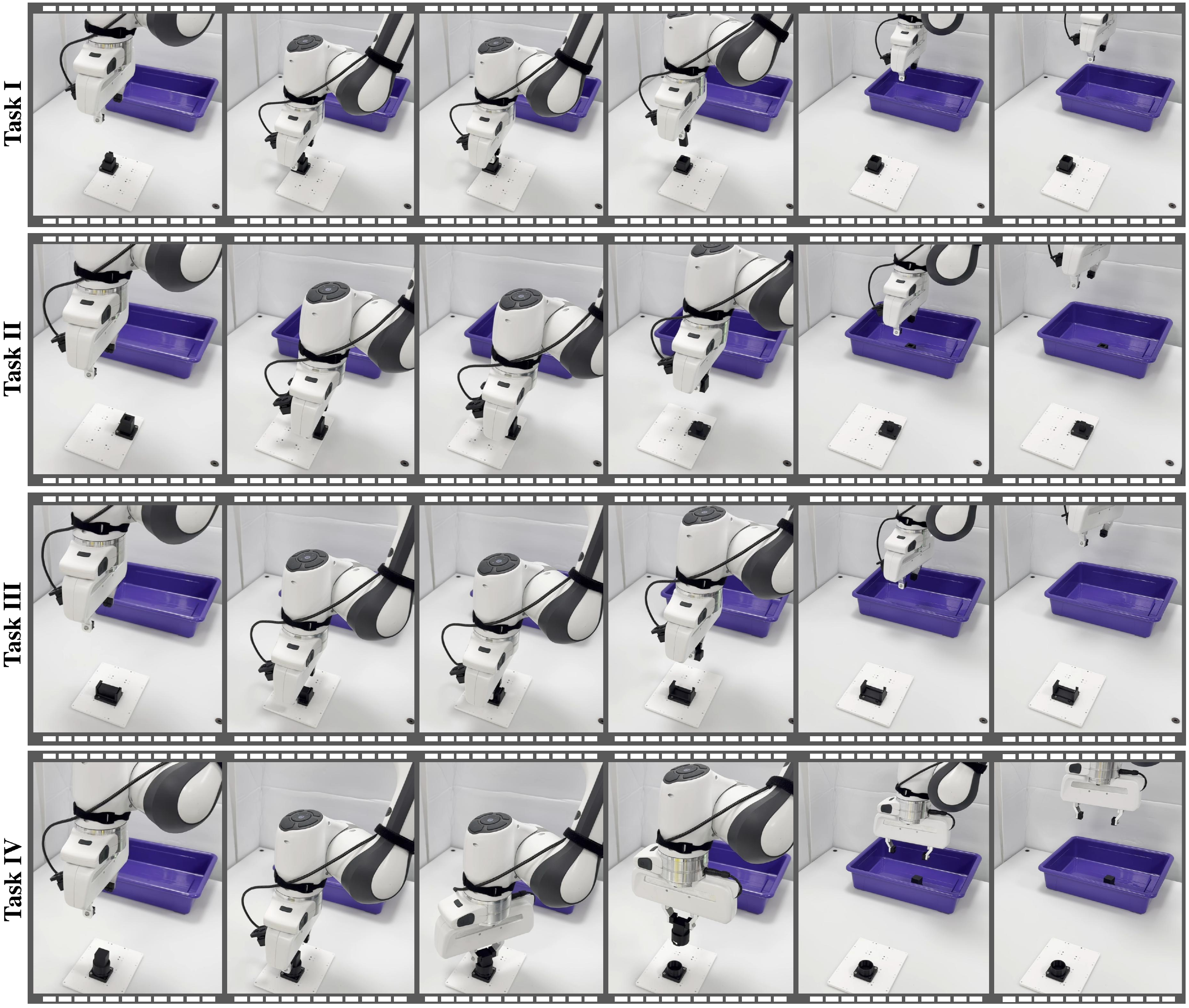}
    \centering
    \captionof{figure}{\small \textbf{\name Deployment in Real-world.} Examples of \name deployment across the considered disassembly tasks, illustrating the policy execution across subsequent frames.}
    \label{fig:realdemo}
\end{figure*}

\subsection{Ablation Studies}
To identify which design choices in \name drive the gains reported in Section~\ref{subsec:mainresults}, we conduct five controlled ablations that each isolate a single factor. the camera view supplying the auxiliary target, the spatial granularity of the latent target, the choice of the frozen encoder $\mathcal{E}_{\mathrm{encoder}}$, the latent-dynamics versus absolute-latent target formulation, and the auxiliary loss weight $\lambda$. Unless otherwise stated, all ablations use LIBERO with patch-level, wrist-camera, latent-dynamics supervision at $\lambda=0.02$ as the reference configuration, matching our main results in Table~\ref{tab:libero-main}.

\vspace{0.2cm}
\noindent
\textbf{Camera View.} We compare three views for the auxiliary target: wrist-mounted, fixed-third person, and a multi-camera variant that concatenates latents from both, using patch-level, latent dynamics ($\mathbf{y}$) supervision at $\lambda = 0.02$. All three views improve over the SmolVLA~\cite{shukor2025smolvla} baseline (\underline{84.1}\%), but the wrist camera yields the largest gain (\underline{88.4}\%) ahead of the multi-camera configuration (\underline{88.1}\%), and the third-person view alone (\underline{86.8}\%). This is due to the fact that the wrist-camera captures the short-range, contact-centric interactions which are the most task-relevant scene dynamics over the planning horizon. The fixed third-person view instead spends capacity on background regions that evolve less informatively as illustrated in Figure~\ref{fig:ablation_cameraview}. This finding is corroborated by our real-world disassembly results in Table~\ref{tab:realworlddeployment}, where wrist camera supervision alone raises success rate from \underline{63.3}\% to \underline{82.5}\%.

\vspace{0.2cm}
\noindent
\textbf{Target Granularity: Patch vs. Mean.} We compare patch-level against mean-pooled latent targets under latent-dynamics ($\mathbf{y}$) supervision at $\lambda=0.02$ across all three camera views. Patch-level supervision outperforms mean pooling in every view: \underline{88.4}\% vs. \underline{86.3}\% on the wrist camera, \underline{88.1}\% vs. \underline{84.4} on the multi-camera configuration, and \underline{86.8}\% vs. \underline{86.1}\% on the third-person view as shown in Figure~\ref{fig:ablation_targetgarnularity}. Mean-pooling discards where the scene changes, whereas patch-level targets retain per-region dynamics, giving the future head a denser and more spatially grounded supervision signal.

\vspace{0.2cm}
\noindent
\textbf{Latent Dynamics vs. Absolute Latent Supervision.} We compare our latent-dynamics target $\mathbf{y}$ against directly supervising absolute future latents $\mathbf{s}$ holding view (wrist), granularity (patch), and $\lambda$ (0.02) fixed. The dynamics formulation reaches \underline{88.4}\% versus \underline{86.7}\% for the absolute variant, with the same ordering, though a narrower margin, on the third-person view (\underline{86.8}\% vs. \underline{86.7}\%). Supervising change rather than absolute content relieves the future head of reconstructing static, task-irrelevant scene content already present in the current observation, concentrating the training signal on action relevant scene evolution as shown in Figure~\ref{fig:ablation_absolutevslatent}.

\vspace{0.2cm}
\noindent
\textbf{Loss Weight $\lambda$.} We also sweep the auxiliary loss weight $\lambda$ using patch-level, wrist-camera, latent-dynamics supervision. Increasing $\lambda$ from 0.005 to 0.02 improves the average success rate from \underline{85.8}\% to \underline{88.4}\% as shown in Figure~\ref{fig:ablation_lambda}. We adopted $\lambda=0.02$ as our default auxiliary loss.

\vspace{0.2cm}
\noindent
\textbf{Encoder Choice: SigLIP~\cite{zhai2023sigmoid} vs. JEPA~\cite{assran2025v}.} \name is agnostic to the frozen encoder $\mathcal{E}_{\mathrm{encoder}}$ used to compute planning-horizon targets. We compare SigLIP~\cite{zhai2023sigmoid}, a vision-language-aligned image encoder, against V-JEPA 2~\cite{assran2025v}, a self-supervised video encoder trained explicitly for dynamics prediction, across camera view, target granularity, and target formulation. On LIBERO, SigLIP is the stronger encoder in every matched configuration we evaluated, e.g., \underline{88.4}\% vs. \underline{86.3}\% at patch, wrist, latent future dynamics, as reported in Table~\ref{tab:ablation_siglip_vs_jepa_libero}. On Meta-World, JEPA is the stronger encoder, reaching \underline{59.8}\% vs. \underline{57.8}\% for SigLIP, as reported in Table~\ref{tab:ablation_siglip_vs_jepa_metaworld}. On real-world disassembly, SigLIP supervision reaches \underline{82.5}\% average success versus \underline{77.5}\% for JEPA-supervision as listed in Table~\ref{tab:realworlddeployment}. Both encoders consistently exceed their respective no-future-supervision baselines across all three benchmarks.

\subsection{Real-world Deployment}
\noindent
\textbf{Experimental Setup.} Our experiment platform consists of a 7-DoF Franka Emika Panda robotic arm equipped with parallel grippers. For visual perception, we employ two RGB cameras: a fixed third-view camera providing a global scene view, and a wrist mounted camera offering close-range, fine-grained observations.

\vspace{0.2cm}
\noindent
\textbf{Disassembly Tasks.} We design a set of contact-rich disassembly tasks where each scenario requires fine-grained adjustments for successful disassembly as shown in Figure~\ref{fig:disassembly_tasks}. The specific disassembly instructions/prompts for each task are listed in Table~\ref{tab:realworlddeployment}.

To train and evaluate \name, we collect a real-world dataset on these disassembly tasks. The dataset contains 100 demonstrations for each task collected by human teleoperation. Visual data consists of RGB images captured from a fixed third-view and wrist-mounted camera. Language data provides prompt instructions for each task as shown in Table~\ref{tab:realworlddeployment}. All modality streams are temporally aligned and synchronized to the frequency of 10 Hz at each timestep.

\vspace{0.2cm}
\noindent
\textbf{Results and Analysis.} Table~\ref{tab:realworlddeployment} reports the success rates across four disassembly tasks among our models (\name, \name (JEPA)), and ACT~\cite{zhao2023learning}, Diffusion Policy~\cite{chi2025diffusion}, and SmolVLA~\cite{shukor2025smolvla} baselines, and Figure~\ref{fig:realdemo} illustrates examples of successful deployment of \name across the disassembly tasks. \name and \name (JEPA) achieve the best overall performance, with an average success rate of \underline{82.5}\% and \underline{77.5}\% outperforming ACT (\underline{35.8}\%), Diffusion Policy (\underline{51.7}\%), and baseline SmolVLA (\underline{63.3}\%). These results demonstrate that reasoning over planning horizon through the wrist-mounted camera provides high quality supervision signal for contact-rich disassembly tasks.

%% file: Conclusions.tex
\section{Conclusions} \label{sec:conclusion}
We presented \name, a training-time planning-horizon supervision framework for VLA policies. \name aligns action-token representations with privileged latent dynamics computed from planning-horizon demonstration frames, then discards the future head and auxiliary encoder at inference. \name improves the success rate consistently across standard manipulation benchmarks and real-world disassembly tasks. Additionally, our ablations demonstrate that target structure, not merely the presence of a future-prediction objective, drives the gain. Patch-level targets outperform mean-pooled targets, wrist-camera supervision outperforms third-person and multi-camera supervision, and supervising latent dynamics outperforms supervising absolute future latents, consistently across configurations we tested.

One of \name's limitations is that it is an auxiliary training objective, not an inference-time planner or world model: it only shapes the policy's representations during fine-tuning but does not perform test-time correction. Extending \name with tactile or force-aware future targets, and object-centric patch supervision are promising directions for future work.